# Demand-Driven Clustering in Relational Domains for Predicting Adverse Drug Events

**Jesse Davis**  JESSE.DAVIS@CS.KULEUVEN.BE
KU Leuven, Celestijnenlaan 200a, Heverlee 3001, Belgium

**Vítor Santos Costa**  VSC@DCC.FC.UP.PT
CRACS INESC-TEC and FCUP Universidade do Porto, Rua do Campo Alegre, 4169-007 PORTO, Portugal

**Peggy Peissig, Michael Caldwell**  {PEISSIG.PEGGY,CALDWELL.MICHAEL}@MARSHFIELDCLINIC.ORG
Marshfield Clinic, 1000 N Oak Ave, Marshfield, WI 54449 USA

**Elizabeth Berg, David Page**  {BERG,PAGE}@BIOSTAT.WISC.EDU
University of Wisconsin - Madison, 1300 University Avenue, Madison, WI 53706 USA

## Abstract

Learning from electronic medical records (EMR) is challenging due to their relational nature and the uncertain dependence between a patient's past and future health status. Statistical relational learning is a natural fit for analyzing EMRs but is less adept at handling their inherent latent structure, such as connections between related medications or diseases. One way to capture the latent structure is via a relational clustering of objects. We propose a novel approach that, instead of pre-clustering the objects, performs a demand-driven clustering during learning. We evaluate our algorithm on three real-world tasks where the goal is to use EMRs to predict whether a patient will have an adverse reaction to a medication. We find that our approach is more accurate than performing no clustering, pre-clustering, and using expert-constructed medical heterarchies.

## 1. Introduction

Statistical relational learning (SRL) (Getoor & Taskar, 2007) focuses on developing learning and reasoning formalisms that combine the benefits of relational representations, such as relational databases or first-order logic, with those of probabilistic, graphical models for handling uncertainty. SRL is especially applicable to domains where it is important to incorporate information from multiple different relations in a learned model and explicitly model uncertainty. One emerging application that meets both criteria is analyzing electronic medical records (EMR). An EMR is a relational database that stores a patient's clinical history: disease diagnoses, procedures, prescriptions, lab results, etc. Using EMRs it is possible to build models to address important medical problems such as predicting which patients are most at risk for having an adverse response to a certain drug. However, EMRs pose challenges due to their relational schemas (i.e., the database contains separate relational tables for diagnoses, prescriptions, labs, etc.), longitudinal nature (e.g., time of diagnosis may be important), and because different patients may have dramatically different numbers of entries in any given table, such as diagnoses or vitals. Furthermore, it is important to model the uncertain, non-deterministic relationships between patients' clinical histories and current and future predictions about their health status.

Latent structure poses a substantial challenge for using machine learning to analyze EMR data. A patient's clinical history records information about specific prescribed medications (e.g., name, dosage, duration) or specific disease diagnoses. It *does not* explicitly mention important connections between different medications or diagnoses, such as which other medications could be prescribed to treat an illness. This information may be necessary to build accurate models. Medical resources provide some relevant information. For example, the ICD9 diagnoses codes are a tree struc-tured hierarchy over a vocabulary of more than 14,000 concepts. Yet, it is impossible for a single, pre-defined hierarchy to capture all the medically-relevant groupings of diseases or medications for a particular prediction task. For example, suppose we are using machine learning to detect if certain antibiotics carry a risk of liver damage, which is a known effect. For any one of these antibiotics, the number of people taking it may be small enough that the association is too weak to meet an interestingness threshold, such as the support threshold in association rule mining. Yet if the algorithm examines all antibiotics, or all antibiotics grouped by their mechanism of action, most drugs in the class do not exhibit the association. Detecting the assocation with the adverse event requires *discovering the right grouping of antibiotics*, and that grouping is unlikely to appear in existing heterarchies.

Addressing this problem requires automatically detecting which diseases or medicines are informative for a specific prediction task and grouping them together. While most state-of-the-art SRL systems are unable to effectively cope with the challenge of latent structure, a few approaches address this problem via a relational clustering of objects and/or relations in a domain (Kemp et al., 2006; Kok & Domingos, 2007; 2008; Sutskever et al., 2010). Intuitively, objects are clustered together if they occur in similar contexts (i.e., participate in the same relations). Some of these approaches aim at discovering relevant structure in the data as opposed to addressing a specific prediction task. Furthermore, a top-down, divisive search is computationally infeasible for complex domains such as EMRs which contain large numbers of objects (e.g., diseases, drugs) (Kemp et al., 2006; Kok & Domingos, 2007). SNE (Kok & Domingos, 2008) is a scalable pre-clustering approach that searches for latent relationships among all objects in a domain. Thus it may miss subtle interactions relevant to the task at hand.

This paper proposes LUCID (Latent Underlying Concept Invention on-Demand), a novel approach that automatically discovers clusters of objects in a relational domain and makes use of the invented clusters in the final learned model. Rather than ignoring the target task and simply pre-clustering objects, LUCID *dynamically clusters objects*, possibly hierarchically, in a *demand-driven* fashion. If it identifies a useful but low coverage regularity in the data, LUCID tries to strengthen it by selecting an object in the regularity and grouping it together with other objects and/or existing groups to expand its coverage. It evaluates if the proposed grouping strengthens the regularity and results in a more accurate learned model. LUCID allows each object to participate in multiple different groupings, as an object may appear in multiple contexts. For example, a drug could be in different groupings related to its mechanism, indications, contraindications, etc.

We motivate and evaluate our proposed approach on the specific task of predicting adverse drug reactions (ADRs) from EMR data. ADRs are the fourth-leading cause of death in the United States and represent a major risk to health, quality-of-life and the economy. The pain reliever Vioxx$^{\text{TM}}$ alone was earning US$2.5 billion per year before it was found to double the risk of heart attack and was pulled from the market while other similar drugs remain on the market. Additionally, accurate predictive models for ADRs are actionable. If a model is found to be accurate in a prospective trial, it could be used to avoid giving a drug to those at highest risk of an ADR. Using three real-world ADR tasks, we demonstrate that the proposed approach produces a more accurate model than using pre-defined medical hierarchies and several other machine learning based approaches. Furthermore, our algorithm uncovered latent structure that a doctor with expertise in our tasks deemed to be interesting and relevant.

## 2. Background

LUCID dynamically constructs clusters that capture latent relationships between different objects in a domain. It does so in the context of the SRL algorithm VISTA (Davis et al., 2007), which combines automated feature construction and model learning into a single process. VISTA uses first-order definite clauses, which can capture relational information, to define (binary) features. These features then become nodes in a Bayesian network.

### 2.1. Datalog

VISTA defines features using the non-recursive Datalog subset of first-order logic.[1] The alphabet consists of three types of symbols: constants, variables, and predicates. *Constants* (e.g., the drug name `Propranolol`), which start with an upper case letter, denote specific objects in the domain. *Variable* symbols (e.g., `disease`), denoted by lower case letters, range over objects in the domain. *Predicate* symbols `P/n`, where $n$ refers to the arity of the predicate and $n \geq 0$, represent relations among objects. An *atom* is $P(t_1, \ldots, t_n)$ where each $t_i$ is a constant or variable. A *literal* is an atom or its negation. A *clause* is a disjunction over a finite set of literals. A *definite clause* is a clause that contains exactly one positive literal. Definite clauses are often written as an implication $B \implies H$, where $B$ is a conjunction of literals called the body and $H$ is a single literal called the head. All variables in a definite clause are assumed to be universally quantified.

---

[1]This subset of first-order logic with a closed-world assumption is equivalent to relational algebra/calculus.

### 2.2. VISTA

VISTA uses definite clauses to define features for the statistical model. Each definite clause becomes a binary feature in the underlying statistical model. The feature receives a value of one for an example if the data about the example satisfies (i.e., proves) the clause and it receives a value of zero otherwise.

VISTA starts by learning a model $M$ over an empty feature set $FS$. This corresponds to a model that predicts the prior probability of the target predicate. Then it repeatedly searches for new features for a fixed number of iterations. In each iteration, VISTA first selects a random seed example and then performs a general-to-specific, breadth-first search through the space of candidate clauses. To guide the search process, it constructs the *bottom clause* by finding all facts that are relevant to the seed example (Muggleton, 1995). VISTA constructs a rule containing just the target attribute, such as ADR(pid), on the right-hand side of the implication. This means that the feature matches all examples. It creates candidate features by adding literals that appear in the bottom clause to the left-hand side of the rule, which makes the feature more specific (i.e., it matches fewer examples). Restricting the candidate literals to those that appear in the bottom clause helps limit the search space while guaranteeing that each generated refinement matches at least one example.

VISTA converts each candidate clause into a feature, $f$, and evaluates $f$ by learning a new model (e.g., the structure of a Bayesian network) that incorporates $f$. In principle, any structure learner could be used, but VISTA typically uses a tree-augmented Naive Bayes model (Friedman et al., 1997). VISTA evaluates each $f$ by comparing the generalization ability of the current model $FS$ versus a model learned over a feature set extended with $f$. VISTA does this by calculating the area under the precision-recall curve (AUC-PR) on a tuning set. AUC-PR is used because relational domains typically have many more negative examples than positive examples, and the AUC-PR ignores the potentially large number of true negative examples.[2] In each iteration, VISTA adds the feature $f'$ to $FS$ that results in the largest improvement in the score of the model. In order to be included in the model, $f'$ must improve the score by a certain percentage-based threshold. This helps control overfitting by pruning relatively weak features that only improve the model score slightly. If no feature improves the model's score, then it simply proceeds to the next iteration.

---

[2] In principle, VISTA can use any evaluation metric to evaluate the quality of the model including (conditional) likelihood, accuracy, ROC analysis, etc.

## 3. LUCID

At a high-level, the key innovation of LUCID occurs when constructing feature definitions. Here, the algorithm has the ability to invent hierarchical clusters that pertain to a subset of the objects (i.e., constants) in the domain. Intuitively, constants that appear in the same grouping share some latent relationship. Discovering and exploiting the latent structure in the feature definitions provides several benefits. First, it allows for more compact feature definitions. Second, by aggregating across groups of objects, it helps identify important features that may not otherwise be deemed relevant by the learning algorithm.

To illustrate the intuition behind LUCID, we use a running example about ADRs to the medication Warfarin$^{\text{TM}}$, which is a blood thinner commonly prescribed to patients at risk of having a stroke. However, Warfarin is known to increase the risk of internal bleeding for some patients. Consider the following feature definition:

$$\text{Drug}(\text{pid}, \text{date1}, \text{Terconazole}) \wedge$$
$$\text{Weight}(\text{pid}, \text{date1}, \text{w}) \wedge \text{w} < 120 \Rightarrow \text{ADR}(\text{pid}) \quad (1)$$

This rule applies only to those patients who satisfy all the conditions on the left hand side of rule. Conditioning on whether a patient has been prescribed Terconazole limits the applicability of this rule. Terconazole is an enzyme inducer, which is a type of medication known to elevate a patient's sensitivity to Warfarin. However, many other drugs in the enzyme inducer class (e.g., Rifampicin and Ketoconazolegive) are frequently prescribed instead of Terconazole, which makes this feature overly specific. A potentially stronger feature would replace Terconazole with an invented concept such as *enzyme inducer* or *Warfarin elevator*.

Yet, these concepts are not explicitly encoded in clinical data. By grouping together related objects, LUCID captures latent structure and is able to learn more general features. For example, we could generalize the previous rule as follows:

$$\text{Cluster1}(\text{did}) \wedge \text{Drug}(\text{pid}, \text{date1}, \text{did}) \wedge$$
$$\text{Weight}(\text{pid}, \text{date1}, \text{w}) \wedge \text{w} < 120 \Rightarrow \text{ADR}(\text{pid}) \quad (2)$$

The definition for `Cluster1`, shown in Cluster definition 3 below, represents latent structure among a group of medicines. Rule (2) is more general than Rule (1), as it is true of any patient that has taken any of the medications assigned to `Cluster1`. Rule (1) is more restrictive because it is only true of patients that have taken Terconazole.

The three key elements of LUCID introduced in the next subsections are: (i) how to represent latent structure, (ii) how to learn the latent structure, and (iii) how the overall algorithm functions.

### 3.1. Representing Latent Structure

LUCID's goal is to capture hierarchical latent structure about specific objects in the domain. Conceptually, clusterings represent the latent structure. LUCID introduces one unary predicate, such as `Cluster1/1`, for each cluster it invents. The predicate is true of any object that is assigned to the cluster it represents. Once the definition has been learned, it can appear in learned rules such as Rule (2). LUCID can assign objects to clusters in two ways.

First, LUCID can assign individual objects to a cluster. This captures that specific constants are interchangeable in some cases. For example, Terconazole, Rifampicin and Ketoconazole are all enzyme inducers, and a doctor could reasonably prescribe any of them. Thus LUCID could invent a new cluster, generically called `Cluster1`, as follows:

$$\text{Cluster1(Terconazole)}$$
$$\text{Cluster1(Rifampicin)} \quad (3)$$
$$\text{Cluster1(Ketoconazole)}$$

These statements simply assign these drugs to `Cluster1`. There is no limit on the number of objects that can be assigned to each invented cluster.

Second, LUCID can reuse previously discovered concepts to represent more high-level, hierarchical structure. It can do so in the following manner:

$$\text{Cluster2(Propranolol)}$$
$$\text{Cluster2(Alpranolol)} \quad (4)$$
$$\text{Cluster1(x)} \Rightarrow \text{Cluster2(x)}$$

Just as before, the first two statements assign specific drugs to `Cluster2`. The key step is the third statement, where all the constants that have been assigned to `Cluster1` are assigned to `Cluster2` as well. Once a proposed grouping has been used in a feature that has been included in the model, it is available for future reuse during the learning procedure. Reusing previously discovered concepts allows the algorithm to automatically explore tradeoffs between fine-grained grouping (e.g., enzyme inducers) and more high-level groupings (e.g., Warfarin elevators) that may be present in the data. Furthermore, it allows the algorithm to build progressively more complex concepts over time.

### 3.2. Learning Latent Structure

The key step in the algorithm is discovering the latent structure. Given a feature definition (e.g., Rule (1)) and a constant to replace with a cluster (e.g., Terconazole), latent structure is learned according to a two-step process. The first step rewrites a feature definition so that it applies to a set of objects, that is an invented cluster, instead of a single, specific object. The second step decides which objects to assign to the newly invented cluster.

**Feature Redefinition** LUCID rewrites the feature definition by replacing the reference to the specific constant with a variable and conjoining an invented latent structure predicate to the end of the rule, as shown in Algorithm 1. For example, Rule (1) would be transformed into Rule (2), where Rule (2) has the variable `did` instead of the constant `Terconazole`, and the feature definition has been extended with the invented predicate `Cluster1(did)`.

---

**Algorithm 1** RewriteRule(Feature $f$, Object `Const`, Feature Set $FS$, Data $D$))

---

Let $B$ be the body of $f$
Let $H$ be the head of $f$
/*Replace `Const` with `newVar` in $B$*/
Substitute($B$, `Const`, `newVar`)
$B^{'} = C_i(\text{newVar}) \land B$
LearnCluster($B^{'} \implies H$, `Const`, $FS, D$)
**return:** $B^{'} \implies H$

---

**Assigning Objects to the Cluster** In order to learn a definition for the invented latent predicate, such as `Cluster1` in Rule (2), LUCID needs to decide which objects to include in the cluster. LUCID employs a bottom-up, data-driven approach for assigning objects to the cluster. The cluster initially contains a single constant and LUCID greedily adds additional constants of the same type to it.

LUCID creates the initial cluster by assigning the replaced constant to it. In the running example, this corresponds to following statement: `Cluster1(Terconazole)`. Next, it tries to identify a set of candidate constants of the same type that could be added to the cluster. Ideally, the candidate set

**Algorithm 2** LearnCluster(Feature $f$, Object `Const`, Feature Set $FS$, Data $D$)

$\quad C_i = \{\text{Const}\}$
$\quad Cand = \text{ConstNearMiss}(f) \cup \{C_j | \text{ cluster } C_j, j < i\}$
$\quad$/*Score the extended feature set*/
$\quad score = \text{AUCPR}(FS \cup f, D)$
$\quad$**repeat**
$\quad\quad$**for all** Candidates $e \in Cand$ **do**
$\quad\quad\quad C_i = C_i \cup \{e\}$
$\quad\quad\quad$/*Score expanded cluster definition*/
$\quad\quad\quad score = \text{AUCPR}(FS \cup f, D)$
$\quad\quad\quad$**if** $newScore > score$ **then**
$\quad\quad\quad\quad Cand = Cand \setminus e$
$\quad\quad\quad\quad score = newScore$
$\quad\quad\quad$**end if**
$\quad\quad$**end for**
$\quad$**until** No addition to $C_i$ improves $score$

would include each object of the same type as well as every previously invented cluster about the type. This is often computationally infeasible due to the large number of objects. For example, when predicting adverse reactions, the data contains information about thousands of drugs and diseases. The central challenge is to identify a small but promising set of candidates to include in a grouping. LUCID restricts its candidate set to objects from "near miss" examples. To illustrate this idea, consider the following two rules:

$$\texttt{Weight}(\texttt{pid}, \texttt{date1}, \texttt{w}) \wedge \texttt{w} < 120 \Rightarrow \texttt{ADR}(\texttt{pid}) \quad (5)$$

$$\texttt{Drug}(\texttt{pid}, \texttt{date1}, \texttt{Terconazole}) \wedge$$
$$\quad \texttt{Weight}(\texttt{pid}, \texttt{date1}, \texttt{w}) \wedge \texttt{w} < 120 \Rightarrow \texttt{ADR}(\texttt{pid}) \quad (6)$$

The second rule, by adding the condition `Drug(pid, date1, Terconazole)`, applies to fewer patients. Some patients may match Rule (5), but not the more specific Rule (6) because they took a similar, but not identical medication. In a sense, Rule (5) provides a context where a drug like Terconazole may be prescribed. Thus, focusing on the medications prescribed to the set of patients that match Rule (5) but not Rule (6) can potentially identify which medications can be prescribed in place of Terconazole. Only considering objects from "near miss" examples has two desirable properties. First, it only adds objects to a cluster that improve a rule's positive minus negative coverage score, meaning the addition is guaranteed to increase recall without harming precision, at least on the training set. Second, all objects with this property are considered, giving the heuristic a completeness property. The candidate set thus includes (i) all previously invented clusters about the same type, and (ii) all constants that appear in examples covered by a rules' immediate predecessor (i.e., Rule (5)) but not the rule itself (i.e., Rule (6)).

Given the candidate set, LUCID tries to extend the definition of the cluster under construction. It adds each candidate, in turn, to the cluster. The benefit of the modified cluster is measured by seeing if the score of the retrained model, which includes the feature that makes use of the extended cluster definition, improves. LUCID greedily selects the single constant that results in the largest improvement in the model's score. This procedure iterates until no addition improves the model's performance or the set of candidate constants is empty. The end result is a cluster definition as illustrated by either Cluster definition (3) or Cluster definition (4).

### 3.3. Overall Algorithmic Structure

Algorithm 3.3 provides an overview of LUCID. It uses the same procedure as VISTA to construct an initial set of candidate features. Next, LUCID considers augmenting each feature definition with an extra, invented predicate by calling the procedure outlined in Subsection 3.2. This is the key difference with VISTA as this results in a larger and much more expressive set of candidate features. However, due to the large number of candidate features, it is prohibitively expensive to consider inventing and incorporating a learned cluster into each feature definition. Therefore, LUCID restricts itself to inventing a latent concepts only for features that meet the following two conditions:

**Condition 1:** The rule under consideration improves the score of the model. This provides initial evidence that the rule is useful, but the algorithm may be able to improve its quality by modeling latent structure. Discarding rules that initially exhibit no improvement dramatically improves the algorithm's efficiency.

**Condition 2:** The rule must contain a constant of a type that the user identified as a candidate for having latent structure. This helps reduce the search space as not all types of constants will exhibit latent structure.

For each candidate feature that meets these two criteria, LUCID attempts to discover latent structure. It invokes the procedure outlined in Subsection 3.2 and adds the feature it constructs, which contains an invented latent predicate, to the set of candidate features. From the expanded candidate feature set, LUCID picks the highest scoring feature $f_{best}$ and adds it to the feature set. If $f_{best}$ contains an invented cluster, then that cluster is made available for reuse in subsequent iterations. If no feature improves the model's score, then LUCID goes to the next iteration. The procedure terminates after running for a fixed number of iterations.

---

**Algorithm 3** LUCID(Data $D$, Maximum Iteration $m$)

$FS = \{\emptyset\}$
**repeat**
  /*Generate Candidate Features*/
  $Cand = $ GenCandidates()
  **for all** ($f \in Cand$) **do**
    **if** ($f$ meets Cond1 AND Cond2) **then**
      /*Select object to replace with cluster*/
      `Const` $\in f$
      $f' = $ RewriteRule($f$, `Const`, $FS, D$)
      $Cand = Cand \cup f'$
    **end if**
  **end for**
  /*Finds highest scoring rule*/
  $f_{best} = $ FindHighScoreRule($Cand$)
  $FS = FS \cup f_{best}$
**until** Reaching iteration $m$
**return:** $FS$

---

## 4. Empirical Evaluation

In this section, we evaluate our proposed approach on three real-world data sets. In all tasks, we are given patients that take a certain medication and the goal is to model the patients that have a related ADR. We compare the following algorithms.

**VISTA** This is the basic VISTA algorithm (Davis et al., 2007). It does not have the ability to learn clusters that capture latent structure.

**Expert+VISTA** In this setting, we augmented each data set with hand-crafted hierarchies for both diagnoses and medications. For diagnoses, we used all the levels of the ICD9 hierarchy. For medications, we use a hierarchy developed by our medical collaborators. We then run VISTA on the augmented data set. Thus, instead of being limited to the specific disease diagnoses or medications recorded for a patient, EXPERT+VISTA can learn rules that exploit information about diseases or medications that appear higher up in the expert defined hierarchies.

**SNE+VISTA** This approach uses the SNE system (Kok & Domingos, 2008) as a pre-clustering step to identify latent structure. SNE is an unsupervised algorithm for automatically clustering to-gether objects that are related to each other. First, SNE analyzes the training data and produces a clustering of the objects in the domain. To exploit SNE's clusters in VISTA, we create one unary predicate for each clustering that is true of every object assigned to that cluster. Second, VISTA is run on the data set which has been augmented with SNE's learned clustering. Thus the learned rules can incorporate the clusters discovered by SNE.

Table 1. Data Set Characteristics.

|  | Selective Cox-2 | Warfarin | ACEi |
|---|---:|---:|---:|
| Pos. examples | 160 | 144 | 102 |
| Neg. examples | 2,134 | 1,440 | 1,020 |
| Unique drugs | 2,590 | 2,316 | 2,044 |
| Unique diagnoses | 7,912 | 8,389 | 7,286 |
| Drug facts | 3,518,467 | 603,503 | 335,065 |
| Diagnoses facts | 3,653,487 | 691,591 | 436,934 |

**LUCID** This is the approach proposed in this paper.

**Expert+LUCID** This gives the proposed approach access to the expert defined hierarchies.

We first describe the data sets we use. Then we present and discuss our experimental results.

### 4.1. Task Descriptions

Our data comes from a large multispecialty clinic that has been using electronic medical records since 1985 and has electronic data back to the early 1960's. We have received institutional review board approval to undertake these studies. For all tasks, we have access to information about observations (e.g., vital signs, family history, etc.), lab test results, disease diagnoses and medications. We only use patient data up to one week before that patient's first prescription of the drug under consideration. This ensures that we are building predictive models only from data generated before a patient is prescribed that drug.

Characteristics of each task can be found in Table 1. We now briefly describe each task. **Selective Cox-2** inhibitors (e.g., Vioxx$^{\text{TM}}$) are a class of pain relief drugs that were found to increase a patients risk of having a a myocardial infarction (MI) (i.e., a heart attack). Angiotensin-converting enzyme inhibitors (**ACEi**) are a class of drugs commonly prescribed to treat high blood-pressure and congestive heart failure. It is known that in some people, ACEi may result in angioedema (a swelling beneath the skin). **Warfarin** is a commonly prescribed blood-thinner that is known to increase the risk of internal bleeding for some individuals. On each task the goal is to distinguish between patients who take the medicine and have an adverse event (i.e., positive examples) and those who do not (i.e., the negative examples).

## 4.2. Methodology and Results

We performed stratified, ten-fold cross-validation for each task. For SNE, we used the default parameter settings. We sub-divided the training data and used five folds for training the model structure and parameters and four folds for tuning. We require that a candidate feature result in at least a 2% improvement to the AUC-PR in order to be considered for acceptance. We set all parameters to be identical for all approaches. The only difference between VISTA and LUCID based approaches is that LUCID can introduce latent structure. Without this ability, the algorithms would construct and evaluate identical candidate feature sets.

Table 2 reports the average AUC-PRs for each of the tasks. LUCID alone outperforms all the non-LUCID approaches on all three tasks. In two of the three, the addition of the ICD9 codes further improves LUCID's performance, while in the other one it degrades LUCID's performance. On the Selective Cox-2 and Warfarin domains, LUCID results in relatively large improvements in AUC-PR, of 12% and 41%, respectively when compared to VISTA. On these two domains, LUCID improves the AUC-PR by 25% and 23% compared to SNE+VISTA. LUCID offers improvements of between 2% and 14% compared to using the expert provided heterarchies. These improvements come with little run-time cost. Across all three tasks, the average run time per fold was approximately 1 hour for VISTA, 8 hours for SNE+VISTA, 1.7 hours for VISTA+Expert, 1.1 hours for LUCID and 1.6 hours for LUCID+Expert. SNE+VISTA is slow because running SNE took between 1 and 16 hours per fold.

There is clearly a benefit to incorporating the latent information about the relationships between medicines and between diseases. In particular, it is beneficial to include the data-driven latent structure and the expert provided heterarchies. Interestingly, the learned structure is *always* more valuable than the expert heterarchies in terms of building an accurate model. This indicates that these resources either lack the relevant groupings of terms or their groupings are not at the right granularity for these prediction tasks. Expert+VISTA achieves the best non-LUCID based result on two of three tasks. Combining LUCID with the expert knowledge is not always useful. The most likely explanation is that this approach has the largest search space and it falls into a local optimum. The utility of pre-clustering prior to learning, represented by SNE+VISTA, is less clear. This approach improves on the baseline on two tasks, but it only does better than the hand-crafted heterarchy on one task. A visual inspection of SNE's learned clustering show that it discovers reasonable concepts from a medical perspective. However, it tends to discover more high-level concepts that are perhaps less useful to the prediction task at hand. In constrast, LUCID's discovery is more task directed and it can also leverage partial feature definitions to detect correlations among objects that arise in the context of a specific rule.

Table 2. Average AUC-PR for each approach. The best results for each task is shown in bold.

|  | Selective Cox-2 | Warfarin | ACEi |
|---|---|---|---|
| Expert+LUCID | **0.438** | 0.187 | **0.302** |
| LUCID | 0.424 | **0.203** | 0.300 |
| Expert+VISTA | 0.416 | 0.177 | 0.272 |
| SNE+VISTA | 0.339 | 0.165 | 0.298 |
| VISTA | 0.377 | 0.143 | 0.286 |

## 4.3. Learned Groupings

Another important evaluation measure is whether LUCID invents interesting and relevant concepts. We presented several invented clusters to a medical doctor with expertise in circulatory diseases. We focus our discussion on structures from the Selective Cox-2 domain. The expert noted a cluster containing the drugs diltiazem, a calcium-channel blocker, and clopidogrel (Plavix$^{TM}$), an antiplatelet agent. These two cardiac drugs are frequently used in acute coronary syndrome, especially after angioplasty. In terms of diseases, the expert highlighted a cluster describing cardiac catheter and coronary angioplasty, which are consistent with acute coronary syndrome and means that a patient is at a high risk of having a MI. Another cluster of interest involved cholecystectomy, a procedure that removes the gall blader, as in females the diagnosis of MI is often confused with gall bladder pain. Finally, the expert remarked on a cluster containing hearing loss as a finding that deserves further investigation.

## 5. Related Work

SRL lies at the intersection of relational learning and graphical model learning. Thus methods for discovering latent structure build on predicate invention in relational learning (e.g., (Muggleton & Buntine, 1988)) and latent variable discovery in propositional graphi-cal models (e.g., (Elidan et al., 2000)). Our approach is closely related to Dietterich and Michalski's (1983) relational learning work on internal disjunction. This operation replaces a constant with a disjunction of several constants. We go beyond this work by allowing reuse of an internal disjunction and most importantly, by explicitly modeling and reasoning about uncertainty in the data and the invented predicates.

Our work is not the first to combine ideas from latent variable discovery and predicate invention to perform cluster-based concept discovery in uncertain, relational domains (Kemp et al., 2006; Kok & Domingos, 2007; 2008; Sutskever et al., 2010; Xu et al., 2006; Popescul & Ungar, 2004). Popescul and Ungar (2004) use a pre-processing step that learns clusterings and then treats cluster membership as an invented feature during learning. In contrast, LUCID uses the learning process to guide cluster construction and it also allows reuse of clusters as part of new clusters. Sutskever et al. (2010) focus only on binary relations, whereas our domains have higher arity relations. Empirically, the SNE system (Kok & Domingos, 2008), which we compare to, outperformed the IRM (Kemp et al., 2006) and MRC (Kok & Domingos, 2007) on a domain of similar complexity and size to those we considered.

## 6. Future Work and Conclusions

We presented LUCID, a novel algorithm that discovers latent structure through a dynamic, demand-driven procedure. During learning, it can invent clusters about objects in the domain and include them in the learned model. We evaluated LUCID by learning models from electronic medical record (EMR) data to predict which patients are most at risk to suffer a given adverse drug reaction (ADR). On all three tasks we investigated, LUCID resulted in improved performance compared to a standard SRL baseline, a pre-clustering based latent structure discovery algorithm, and using expert-constructed medical heterarchies. Additionally, it produced meaningful latent structure. Important directions for further research include applications to other ADRs, other tasks in learning from EMRs, and other types of relational databases.

## Acknowledgements

We thank Daniel Lowd, Maurice Bruynooghe and the reviewers for their helpful feedback. JD is partially supported by the Research Fund K.U.Leuven (CREA/11/015 and OT/11/051), EU FP7 Marie Curie Career Integration Grant (#294068) and FWO-Vlaanderen (G.0356.12). VSC is funded by ERDF through Programme COMPETE and by the Portuguese Government through FCT Foundation for Science and Technology projects LEAP (PTDC/EIA-CCO/112158/2009) and ADE (PTDC/EIA-EIA/121686/2010). MC, PP, EB and DP gratefully acknowledge the support of NIGMS grant R01GM097618-01.

## References

Davis, J., Ong, I., Struyf, J., Burnside, E., Page, D., and Costa, V. Santos. Change of representation for statistical relational learning. In *Proc. of the 20th International Joint Conference on Artificial Intelligence*, pp. 2719–2726, 2007.

Dietterich, T. G. and Michalski, R. S. A comparative review of selected methods for learning from examples. In *Machine Learning: An Artificial Intelligence Approach*, pp. 41–81. TIOGA Publishing Co., 1983.

Elidan, G., Lotner, N., Friedman, N., and Koller, D. Discovering hidden variables: A structure-based approach. In *Neural Information Processing Systems 13*, pp. 479–485, 2000.

Friedman, N., Geiger, D., and Goldszmidt, M. Bayesian networks classifiers. *Machine Learning*, 29:131–163, 1997.

Getoor, L. and Taskar, B. (eds.). *An Introduction to Statistical Relational Learning*. MIT Press, 2007.

Kemp, C., Tenenbaum, J., Griffiths, T., Yamada, T., and Ueda, N. Learning systems of concepts with an infinite relational model. In *Proc. of the 21st National Conference on Artificial Intelligence*, 2006.

Kok, S. and Domingos, P. Statistical predicate invention. In *Proc. of the 24th International Conference on Machine Learning*, pp. 433–440, 2007.

Kok, S. and Domingos, P. Extracting semantic networks from text via relational clustering. In *Proc. of the European Conference on Machine Learning and Knowledge Discovery in Databases*, pp. 624–639, 2008.

Muggleton, S. Inverse entailment and Progol. *New Generation Computing*, 13:245–286, 1995.

Muggleton, S. and Buntine, W. Machine invention of first-order predicates by inverting resolution. In *Proc. of the 5th International Conference on Machine Learning*, pp. 339–352, 1988.

Popescul, A. and Ungar, L. Cluster-based concept invention for statistical relational learning. In *Proc. of the 10th ACM International Conference on Knowledge Discovery and Data Mining*, pp. 665–670, 2004.

Sutskever, I., Salakhutdinov, R., and Tenenbaum, J. Modelling relational data using Bayesian clustered tensor factorization. In *Neural Information Processing Systems 23*, 2010.

Xu, Z., Tresp, V., Yu, K., and Kriegel, H-P. Infinite hidden relational models. In *Proc. of the 22nd Conference on Uncertainty in Artificial Intelligence*, 2006.